\def\year{2020}\relax
\documentclass[letterpaper]{article} 
\usepackage{aaai20}  
\usepackage{times}  
\usepackage{helvet} 
\usepackage{courier}  
\usepackage[hyphens]{url}  
\usepackage{graphicx} 
\usepackage{makecell}
\usepackage{hyperref}       
\usepackage{url}
\usepackage{booktabs}       
\usepackage{amsfonts}
\usepackage{nicefrac}      
\usepackage{microtype}   
\usepackage{lipsum}
\usepackage{verbatim}
\usepackage{multirow}
\usepackage{multicol}
\usepackage{color}
\usepackage{latexsym}
\usepackage{graphicx}
\usepackage{xcolor}
\usepackage{colortbl,booktabs}
\usepackage{tabu}
\usepackage{float}
\usepackage{amsmath}
\usepackage{diagbox}
\usepackage{graphicx}  
\title{ERNIE 2.0: A Continual Pre-Training Framework for Language Understanding}
\author{Yu Sun, Shuohuan Wang, Yukun Li, Shikun Feng, Hao Tian, Hua Wu, Haifeng Wang \\
Baidu Inc., Beijing, China \\
\{sunyu02, wangshuohuan, tianhao, wu\_hua,wanghaifeng\}@baidu.com
}
 
\begin{document}
\maketitle
\begin{abstract}
Recently pre-trained models have achieved state-of-the-art results in various language understanding tasks.
Current pre-training procedures usually focus on training the model with several simple tasks to grasp the co-occurrence of words or sentences. 
However, besides co-occurring information, there exists other valuable lexical, syntactic and semantic information in training corpora, such as named entities, semantic closeness and discourse relations. 
In order to extract the lexical, syntactic and semantic information from training corpora, we propose a continual pre-training framework named ERNIE 2.0 which incrementally builds pre-training tasks and then learn pre-trained models on these constructed tasks via continual multi-task learning. 
Based on this framework, we construct several tasks and train the ERNIE 2.0 model to capture lexical, syntactic and semantic aspects of information in the training data.
Experimental results demonstrate that ERNIE 2.0 model outperforms BERT and XLNet on 16 tasks including English tasks on GLUE benchmarks and several similar tasks in Chinese.
The source codes and pre-trained models have been released at https://github.com/PaddlePaddle/ERNIE.
\end{abstract}

\section{Introduction}
Pre-trained language representations such as ELMo\cite{peters2018deep}, OpenAI GPT\cite{radford2018improving}, BERT \cite{devlin2018bert}, ERNIE 1.0 \cite{sun2019ernie}\footnote{In order to distinguish ERNIE 2.0 framework and the ERNIE model, the latter is referred to as ERNIE 1.0.\cite{sun2019ernie}} and XLNet\cite{yang2019xlnet} have been proven to be effective for improving the performances of various natural language understanding tasks including sentiment classification \cite{socher2013recursive}, natural language inference \cite{bowman2015large}, named entity recognition \cite{sang2003introduction} and so on.

Generally the pre-training of models often train the model based on the co-occurrence of words and sentences. While in fact, there are other lexical, syntactic and semantic information worth examining in training corpora other than co-occurrence. 
For example, named entities like person names, location names, and organization names, may contain conceptual information. Information like sentence order and sentence proximity enables the models to learn structure-aware representations. 
And semantic similarity at the document level or discourse relations among sentences allow the models to learn semantic-aware representations. 
In order to discover all valuable information in training corpora, be it lexical, syntactic or semantic representations, we propose a continual pre-training framework named ERNIE 2.0 which could incrementally build and train a large variety of pre-training tasks through continual multi-task learning.

Our ERNIE framework supports the introduction of various customized tasks continually, which is realized through continual multi-task learning. When given one or more new tasks, the continual multi-task learning method simultaneously trains the newly-introduced tasks together with the original tasks in an efficient way, without forgetting previously learned knowledge. In this way, our framework can incrementally train the distributed representations based on the previously trained parameters that it grasped. Moreover, in this framework, all the tasks share the same encoding networks, thus making the encoding of lexical, syntactic and semantic information across different tasks possible.

In summary, our contributions are as follows:
\begin{itemize}
    \item We propose a continual pre-training framework ERNIE 2.0, which efficiently supports customized training tasks and continual multi-task learning in an incremental way.
    \item We construct three kinds of unsupervised language processing tasks to verify the effectiveness of the proposed framework. Experimental results demonstrate that ERNIE 2.0 achieves significant improvements over BERT and XLNet on 16 tasks including English GLUE benchmarks and several Chinese tasks. 
    \item Our fine-tuning code of ERNIE 2.0 and models pre-trained on English corpora are available at \url{https://github.com/PaddlePaddle/ERNIE}.
\end{itemize}

\begin{figure*} 
\centerline
{\includegraphics[width=1.9\columnwidth]{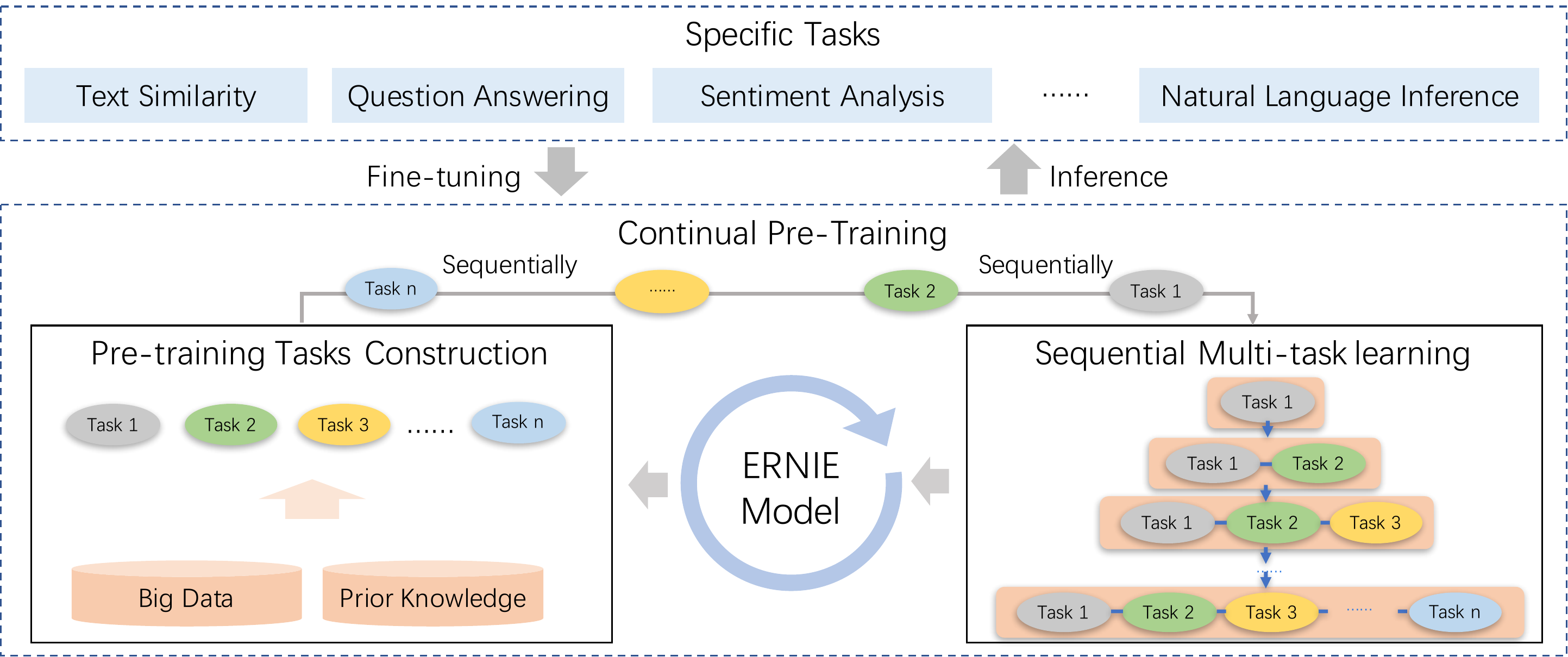}}
\caption{The framework of ERNIE 2.0, where the pre-training tasks can be incrementally constructed, the models are pre-trained through continual multi-task learning, and the pre-trained model is fine-tuned to adapt to various language understanding tasks.}
\label{the framework of ERNIE}
\end{figure*}

\section{Related Work}
\subsection{Unsupervised Learning for Language Representation}
It is effective to learn general language representation by pre-training a language model with a large amount of unannotated data. Traditional methods usually focus on context-independent word embedding. Methods such as Word2Vec \cite{mikolov2013efficient} and GloVe \cite{pennington2014glove} learn fixed word embeddings based on word co-occurring information on large corpora.

Recently, several studies centered on contextualized language representations have been proposed and context-dependent language representations have shown state-of-the-art results in various natural language processing tasks.
ELMo \cite{peters2018deep} proposes to extract context-sensitive features
from a language model. 
OpenAI GPT \cite{radford2018improving} enhances the context-sensitive embedding
by adjusting the Transformer \cite{vaswani2017attention}.
BERT \cite{devlin2018bert}, however, adopts a masked language model while adding a next sentence prediction task into the pre-training.
XLM \cite{lample2019cross} integrates two methods to learn cross-lingual language models, namely the unsupervised method that relies only on monolingual data and the supervised method that leverages parallel bilingual data. 
MT-DNN \cite{liu2019multi} achieves a better result through learning several supervised tasks in GLUE\cite{wang2018glue} together based on the pre-trained model, which eventually leads to improvements on other supervised tasks that are not learned in the stage of multi-task supervised fine-tuning. 
XLNet \cite{yang2019xlnet} uses Transformer-XL \cite{dai2019transformer} and proposes a generalized autoregressive pre-training method that learns bidirectional contexts by maximizing the expected likelihood over all permutations of the factorization order.

\begin{figure} 
\centerline
{\includegraphics[width=0.95\columnwidth]{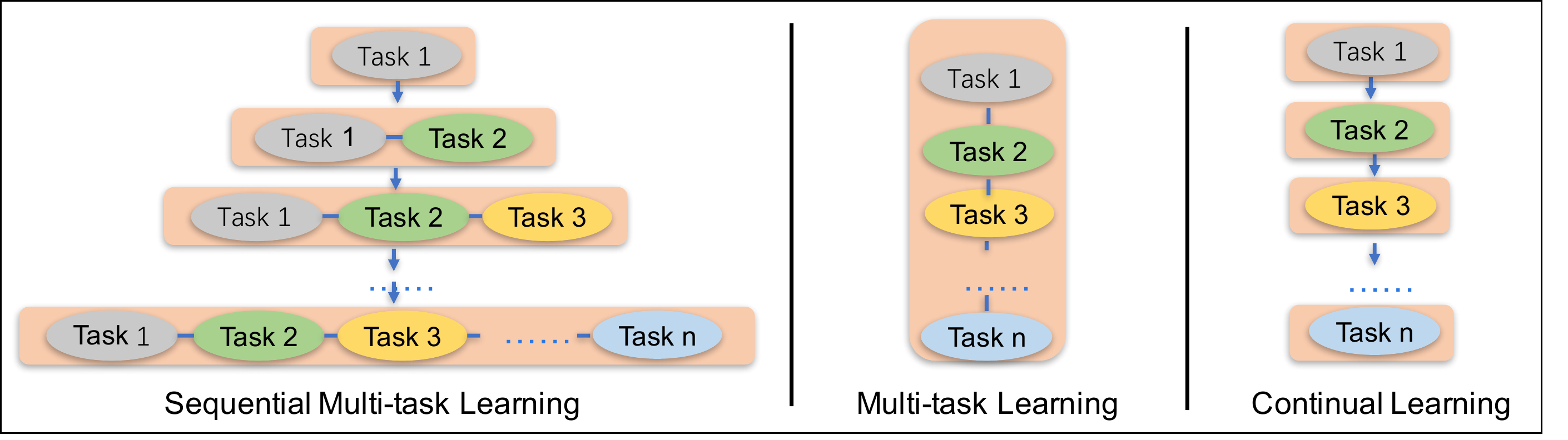}}
\caption{The different methods of continual pre-training.}
\label{diff_method}
\end{figure}

\subsection{Continual Learning}
Continual learning\cite{parisi2019continual,chen2018lifelong} aims to train the model with several tasks in sequence so that it remembers the previously-learned tasks when learning the new ones. These methods are inspired by the learning process of humans, as humans are capable of continuously accumulating the information acquired by study or experience to efficiently develop new skills. With continual learning, the model should be able to performs well on new tasks thanks to the knowledge acquired during previous training.
\begin{figure*} 
\centerline
{\includegraphics[width=1.9\columnwidth]{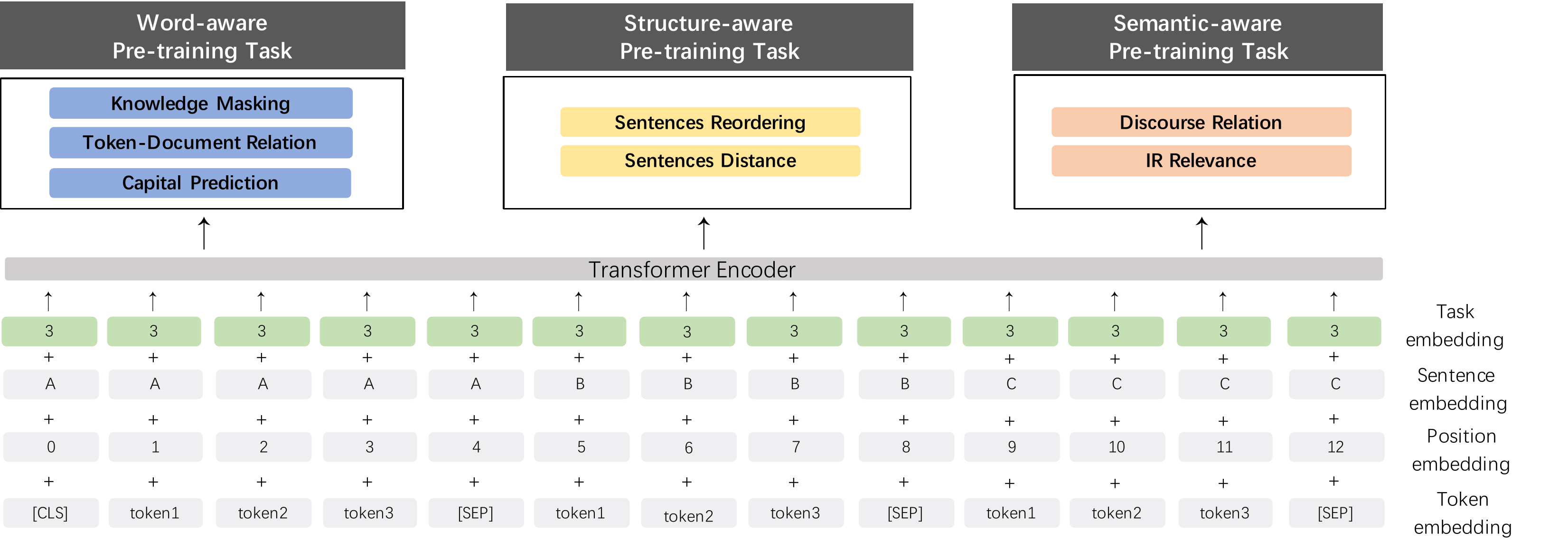}}
\caption{The structure of the ERNIE 2.0 model. The input embedding contains the token embedding, the sentence embedding, the position embedding and the task embedding. Seven pre-training tasks belonging to different kinds are constructed in the ERNIE 2.0 model.}
\label{model_structure}
\end{figure*}

\begin{figure} 
\centerline
{\includegraphics[width=0.95\columnwidth]{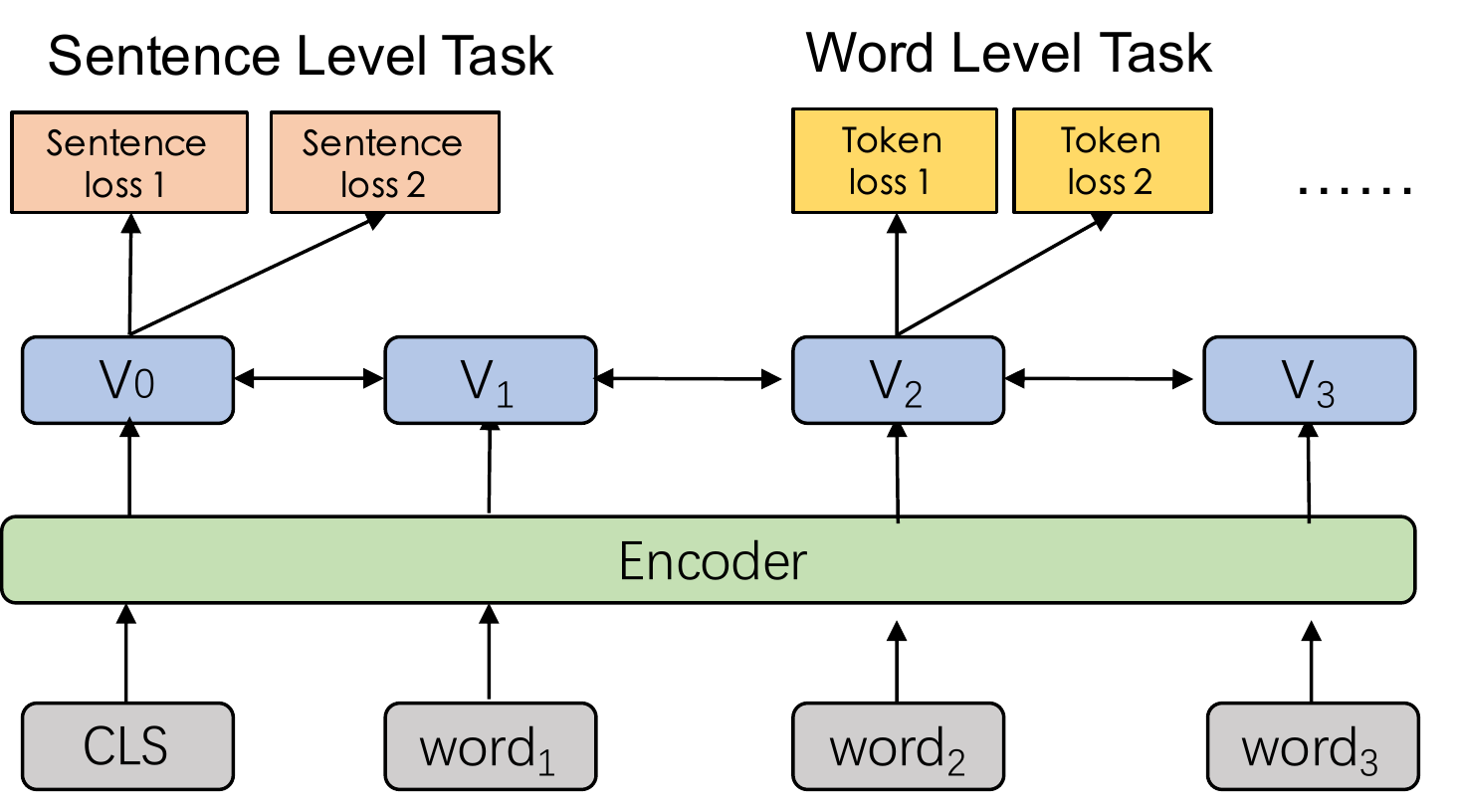}}
\caption{The architecture of multi-task learning in the ERNIE 2.0 framework, in which the encoder can be recurrent neural networks or a deep transformer.}
\label{ernie_loss}
\end{figure}

\section{The ERNIE 2.0 Framework}
As shown in Figure \ref{the framework of ERNIE}, the ERNIE 2.0 framework is built based on an widely-used architecture of pre-training and fine-tuning. 
ERNIE 2.0 differs from the previous pre-training ones in that, instead of training with a small number of pre-training objectives, it could constantly introduce a large variety of pre-training tasks to help the model efficiently learn the lexical, syntactic and semantic representations. 
Based on this, ERNIE 2.0 framework keeps updating the pre-trained model with continual multi-task learning. 
During fine-tuning, the ERNIE model is first initialized with the pre-trained parameters, and would be later fine-tuned using data from specific tasks.

\subsection{Continual Pre-training}
The process of continual pre-training contains two steps. Firstly, We continually construct unsupervised pre-training tasks with big data and prior knowledge involved. Secondly, We incrementally update the ERNIE model via continual multi-task learning. 
\subsubsection{Pre-training Tasks Construction}
We can construct different kinds of tasks at each time, including word-aware tasks, structure-aware tasks and semantic-aware tasks\footnote{For the detailed information of these tasks, please refer to the next section.}. All of these pre-training tasks rely on self-supervised or weak-supervised signals that could be obtained from massive data without human annotation. Prior knowledge such as named entities, phrases and discourse relations is used to generate labels from large-scale data.

\subsubsection{Continual Multi-task Learning}
The ERNIE 2.0 framework aims to learn lexical, syntactic and semantic information from a number of different tasks. Thus there are two main challenges to overcome. The first is how to train the tasks in a continual way without forgetting the knowledge learned before. The second is how to pre-train these tasks in an efficient way. We propose a continual multi-task learning method to tackle with these two problems. Whenever a new task comes, the continual multi-task learning method first uses the previously learned parameters to initialize the model, and then train the newly-introduced task together with the original tasks simultaneously. This will make sure that the learned parameters encodes the previously-learned knowledge. One left problem is how to make it trained more efficiently. We solve this problem by allocating each task N training iterations. Our framework needs to automatically assign these N iterations for each task to different stages of training. In this way, we can guarantee the efficiency of our method without forgetting the previously trained knowledge \footnote{For more details, please refer to Table \ref{continual_learning} in the experiment section.}. 



Figure \ref{diff_method} shows the difference among our method, multi-task learning from scratch and previous continual learning.  
Although multi-task learning from scratch could train multiple tasks at the same time, it is necessary that all customized pre-training tasks are prepared before the training could proceed. So this method takes as much time as continual learning does, if not more. Traditional continual learning method trains the model with only one task at each stage with the demerit that it may forget the previously learned knowledge. 
\begin{table*}[htpb]
  \centering
\small
\begin{center}
{
\begin{tabular}{@{}r|c|c|c|c|c|c|c@{}}
  \hline
  \hline
   \multirow{2}{*}{\diagbox{Corpus}{Task}} & \multicolumn{3}{c|}{Token-Level Loss} & \multicolumn{4}{c}{Sentence-Level Loss}\\ \cline{2-8}
   
   & \makecell{Knowledge \\ Masking} & \makecell{Capital \\ Prediction} & \makecell{Token-Document \\ Relation} & \makecell{Sentence \\ Reordering} & \makecell{Sentence \\ Distance} & \makecell{Discourse \\ Relation} & \makecell{IR \\ Relevance} \\ \hline
  Encyclopedia  & $\checkmark$ & $\checkmark$ & $\checkmark$ & $\checkmark$ & $\checkmark$ & $\times$ &  $\times$  \\ 
  BookCorpus  & $\checkmark$ & $\checkmark$ & $\checkmark$ & $\checkmark$ & $\checkmark$ & $\times$ & $\times$ \\ 
  News & $\checkmark$ & $\checkmark$ & $\checkmark$ & $\checkmark$ & $\checkmark$ & $\times$ & $\times$ \\ \
  Dialog & $\checkmark$ & $\checkmark$ & $\checkmark$ & $\checkmark$ & $\checkmark$ & $\times$ & $\times$  \\ 
  IR Relevance Data & $\times$ & $\times$ & $\times$ & $\times$ & $\times$ &  $\times$ & $\checkmark$ \\
  Discourse Relation Data & $\times$ & $\times$ & $\times$ & $\times$  & $\times$ & $\checkmark$ & $\times$ \\
\hline
\hline
\end{tabular}
} 
\end{center}
\caption{The Relationship between pre-training task and pre-training dataset. We use different pre-training dataset to construct different tasks. A type of pre-trained dataset can correspond to multiple pre-training tasks. }
\label{task_and_dataset}
\end{table*}



As shown in Figure \ref{ernie_loss}, the architecture of our continual multi-task learning in each stage contains a series of shared text encoding layers to encode contextual information, which can be customized by using recurrent neural networks or a deep Transformer consisting of stacked self-attention layers\cite{vaswani2017attention}.
The parameters of the encoder can be updated across all learning tasks. 
There are two kinds of loss functions in our framework. 
One is the sentence-level loss and the other one is the token-level loss, which are similar to the loss functions of BERT.
Each pre-training task has its own loss function. 
During pre-training, one sentence-level loss function can be combined with multiple token-level loss functions to continually update the model.

\subsection{Fine-tuning for Application Tasks}
By virtue of fine-tuning with task-specific supervised data, the pre-trained model can be adapted to different language understanding tasks, such as question answering, natural language inference, and semantic similarity. Each downstream task has its own fine-tuned models after being fine-tuned.

\section{ERNIE 2.0 Model}
In order to verify the effectiveness of the framework, we construct three different kinds of unsupervised language processing tasks and develop a pre-trained model called ERNIE 2.0 model. In this section we introduce the implementation of the model in the proposed framework.

\subsection{Model Structure}
\subsubsection{Transformer Encoder} \,\,
The model uses a multi-layer Transformer\cite{vaswani2017attention} as the basic encoder like other pre-training models such as GPT\cite{radford2018improving}, BERT\cite{devlin2018bert} and XLM\cite{lample2019cross}. The transformer can capture the contextual information for each token in the sequence via self-attention, and generate a sequence of contextual embeddings. Given a sequence, the special classification embedding [CLS] is added to the first place of the sequence. Furthermore, the symbol of [SEP] is added as the separator in the intervals of the segments for the multiple input segment tasks.

\subsubsection{Task Embedding} \,\,
The model feeds task embedding to represent the characteristic of different tasks. We represent different tasks with an id ranging from 0 to N. Each task id is assigned to one unique task embedding. The corresponding token, segment, position and task embedding are taken as the input of the model. We can use any task id to initialize our model in the fine-tuning process. The model structure is shown in Figure \ref{model_structure}.

\subsection{Pre-training Tasks}
We construct three different kinds of tasks to capture different aspects of information in the training corpora.
The word-aware tasks enable the model to capture the lexical information, the structure-aware tasks enable the model capture the syntactic information of the corpus and the semantic-aware tasks aims to learn semantic information. 

\subsection{Word-aware Pre-training Tasks}

\subsubsection{Knowledge Masking Task} \,\, 
ERNIE 1.0\cite{sun2019ernie} proposed an effective strategy to enhance representation through knowledge integration. It introduced phrase masking and named entity masking and predicts the whole masked phrases and named entities to help the model learn the dependency information in both local contexts and global contexts. We use this task to train an initial version of the model.

\subsubsection{Capitalization Prediction Task} \,\, 
Capitalized words usually have certain specific semantic information compared to other words in sentences. The cased model has some advantages in tasks like named entity recognition while the uncased model is more suitable for some other tasks. To combine the advantages of both models, we add a task to predict whether the word is capitalized or not. 

\subsubsection{Token-Document Relation Prediction Task}  \,\,
This task predicts whether the token in a segment appears in other segments of the original document. Empirically, the words that appear in many parts of a document are usually commonly-used words or relevant with the main topics of the document. Therefore, through identifying the frequently-occurring words of a document appearing in the segment, the task can enable the ability of a model to capture the key words of the document to some extent.

\subsection{Structure-aware Pre-training Tasks}

\subsubsection{Sentence Reordering Task} \,\, 
This task aims to learn the relationships among sentences. During the pre-training process of this task, a given paragraph is randomly split into 1 to m segments and then all of the combinations are shuffled by a random permuted order. We let the pre-trained model to reorganize these permuted segments, modeled as a k-class classification problem where \begin{math} k=\sum_{n=1}^{m} n!\end{math}. Empirically, the sentences reordering task can enable the pre-trained model to learn relationships among sentences in a document.

\subsubsection{Sentence Distance Task} \,\, 
We also construct a pre-training task to learn the sentence distance using document-level information. This task is modeled as a 3-class classification problem. "0" represents that the two sentences are adjacent in the same document, "1" represent that the two sentences are in the same document, but not adjacent, and "2" represents that the two sentences are from two different documents.

\subsection{Semantic-aware Pre-training Tasks}

\subsubsection{Discourse Relation Task} \,\, 
Beside the distance task mentioned above, we introduce a task to predict the semantic or rhetorical relation between two sentences. We use the data built by Sileo et.al\cite{sileo2019mining} to train a pre-trained model for English tasks. Following the method in Sileo et.al\cite{sileo2019mining}, we also automatically construct a Chinese dataset for pre-training.

\subsubsection{IR Relevance Task}  \,\, 
We build a pre-training task to learn the short text relevance in information retrieval. It is a 3-class classification task which predicts the relationship between a query and a title. We take the query as the first sentence and the title as the second sentence. The search log data from a commercial search engine is used as our pre-training data.
There are three kinds of labels in this task. The query and title pairs that are labelled as " 0" stand for strong relevance, which means that the title is clicked by the users after they input the query. Those labelled as "1" represent weak relevance, which implies that when the query is input by the users, these titles appear in the search results but failed to be clicked by users. The label "2" means that the query and title are completely irrelevant and random in terms of semantic information.

\section{Experiments}
We compare the performance of ERNIE 2.0 with the state-of-the-art pre-training models. For English tasks, we compare our results with BERT \cite{devlin2018bert} and XLNet \cite{yang2019xlnet} on GLUE. For Chinese tasks, we compare the results with that of BERT \cite{devlin2018bert} and the previous ERNIE 1.0 \cite{sun2019ernie} model on several Chinese datasets. Moreover, we will compare  our method with multi-task learning and traditional continual learning.
\begin{table}[htpb]
  \centering
\small
\begin{center}
{
\begin{tabular}{@{}rcc@{}}
  \hline
  \hline
  Corpus Type     & English(\#tokens) & Chinese(\#tokens) \\ \hline
  Encyclopedia & 2021M & 7378M  \\ 
  BookCorpus  & 805M &  - \\ 
  News  & - &  1478M \\ \
  Dialog  & 4908M & 522M  \\ 
  IR Relevance Data & - & 4500M \\
  Discourse Relation Data & 171M & 1110M \\
\hline
\hline
\end{tabular}
} 
\end{center}
\caption{The size of pre-training datasets. }
\label{training_dataset}
\end{table}
\subsection{Pre-training and Implementation}
\subsubsection{Pre-training Data}
Similar to that of BERT, some data in the English corpus are crawled from Wikipedia and BookCorpus. Besides we also collect some data from Reddit and use the Discovery data \cite{sileo2019mining} as our discourse relation data. For the Chinese corpus, we collect a variety of data, such as encyclopedia, news, dialogue, information retrieval and discourse relation data from a search engine.
The details of the pre-training data are shown in Table \ref{training_dataset}. The relationship between pre-training task and pre-training dataset is shown in Table \ref{task_and_dataset}.

\subsubsection{Pre-training Settings}
To compare with BERT\cite{devlin2018bert}, We use the same model settings of transformer as BERT. 
The base model contains 12 layers, 12 self-attention heads and 768-dimensional of hidden size while the large model contains 24 layers, 16 self-attention heads and 1024-dimensional of hidden size. The model settings of XLNet \cite{yang2019xlnet} are same as BERT. 
\begin{table}[htbp]
\small
\begin{center}
\resizebox{0.45\textwidth}{!}
{
\begin{tabular}{@{}c|ccc|ccc@{}}
  \hline \hline 
   \multirow{2}{*}{Task}      &  \multicolumn{3}{c|}{\textit{BASE}} & \multicolumn{3}{c}{\textit{LARGE}} \\ \cline{2-7}
                &  Epoch & \makecell{Learning \\ Rate}  & \makecell{Batch \\ Size} &  Epoch & \makecell{Learning \\ Rate} & \makecell{Batch \\ Size} \\ \hline 
  CoLA  & 3 & 3e-5 & 64  & 5 & 3e-5 & 32 \\
  SST-2  & 4 & 2e-5 & 256 & 4 & 2e-5 & 64  \\
  MRPC  & 4 & 3e-5 & 32 & 4 & 3e-5 & 16 \\
  STS-B  & 3 & 5e-5 & 128 & 3 & 5e-5 & 128 \\
  QQP  & 3 & 3e-5 & 256  & 3 & 5e-5 & 256  \\
  MNLI  & 3 & 3e-5 & 512 & 3 & 3e-5 & 256 \\
  QNLI  & 4 & 2e-5 & 256 & 4 & 2e-5 & 256  \\
  RTE  & 4 & 2e-5 & 4 & 5 & 3e-5 & 16  \\
  WNLI  & 4 & 2e-5 & 8 & 4 & 2e-5 & 8   \\
\hline \hline
\end{tabular}
} 
\end{center}
\caption{The Experiment settings for GLUE dataset}
\label{GLUE_finetune_setting}
\end{table}
\begin{table}[htbp]
\small
\begin{center}
\resizebox{0.45\textwidth}{!}
{
\begin{tabular}{@{}c|ccc|ccc@{}}
  \hline \hline
   \multirow{2}{*}{Task}      &  \multicolumn{3}{c|}{\textit{BASE}} & \multicolumn{3}{c}{\textit{LARGE}} \\ \cline{2-7}
                &  Epoch & \makecell{Learning \\ Rate} & \makecell{Batch \\ Size} &  Epoch & \makecell{Learning \\Rate}  & \makecell{Batch \\ Size} \\ \hline 
  CMRC 2018  & 2 & 3e-5 & 64 & 2 & 3e-5 & 64 \\
  DRCD  & 2 & 5e-5 & 64 & 2 & 3e-5 & 64 \\
  DuReader  & 2 & 5e-5 & 64 & 2 & 2e-5 & 64 \\
  MSRA-NER  & 6 & 5e-5 & 16    & 6 & 1e-5 & 16 \\
  XNLI  & 3 & 1e-4 &  512 & 3 & 4e-5 & 512  \\
  ChnSentiCorp  & 10 & 5e-5 & 24  & 10 & 1e-5 & 24  \\
  LCQMC  & 3 & 2e-5 & 32  & 3 & 5e-6 & 32  \\
  BQ Corpus  & 3 & 3e-5 & 64  & 3 & 1.5e-5 & 64  \\
  NLPCC-DBQA  & 3 & 2e-5 & 64 & 3 & 1e-5 & 64 \\
\hline \hline
\end{tabular}

} 
\end{center}
\caption{The Experiment Settings for Chinese datasets}
\label{chinese_finetune_setting}
\end{table}

ERNIE 2.0 is trained on 48 NVidia v100 GPU cards for the base model and 64 NVidia v100 GPU cards for the large model in both English and Chinese. The ERNIE 2.0 framework is implemented on PaddlePaddle, which is an end-to-end open source deep learning platform developed by Baidu. We use Adam optimizer that parameters of which are fixed to $\beta_1=0.9$, $\beta_2=0.98$, with a batch size of 393216 tokens. The learning rate is set as 5e-5 for English model and 1.28e-4 for Chinese model. It is scheduled by decay scheme noam \cite{vaswani2017attention} with warmup over the first 4,000 steps for every pre-training task. By virtue of float16 operations, we manage to accelerate the training and reduce the memory usage of our models. Each of the pre-training tasks is trained until the metrics of pre-training tasks converge. 

\begin{table*}[htbp]
  \centering
  \small
\begin{tabular}{c|cc|ccc|cc}
\hline \hline
\multirow{3}{*}{Task(Metrics)}       & \multicolumn{2}{c|}{\textit{BASE model}}   & \multicolumn{5}{c}{\textit{LARGE model}}                                                                                                        \\ \cline{2-8} 
                                     & \multicolumn{2}{c|}{Test}                  & \multicolumn{3}{c|}{Dev}                                                              & \multicolumn{2}{c}{Test}                                 \\ \cline{2-8} 
                                     & BERT      & \multicolumn{1}{c|}{ERNIE 2.0} & \multicolumn{1}{c}{BERT} & \multicolumn{1}{c}{XLNet} & \multicolumn{1}{c|}{ERNIE 2.0} & \multicolumn{1}{c}{BERT} & \multicolumn{1}{c}{ERNIE 2.0} \\ \hline
CoLA (Matthew Corr.)                 & 52.1      & \textbf{55.2}                  & 60.6                     & 63.6                      & \textbf{65.4}                  & 60.5                     & \textbf{63.5}                 \\
SST-2 (Accuracy)                     & 93.5      & \textbf{95.0}                  & 93.2                     & 95.6                      & \textbf{96.0}                  & 94.9                     & \textbf{95.6}                 \\
MRPC (Accurary/F1)                   & 84.8/88.9 & \textbf{86.1/89.9}             & 88.0/-                   & 89.2/-                    & \textbf{89.7/-}                & 85.4/89.3                & \textbf{87.4/90.2}            \\
STS-B (Pearson Corr./Spearman Corr.) & 87.1/85.8 & \textbf{87.6/86.5}             & 90.0/-                   & 91.8/-                    & \textbf{92.3/-}                & 87.6/86.5                & \textbf{91.2/90.6}            \\
QQP (Accuracy/F1)                    & 89.2/71.2 & \textbf{89.8/73.2}             & 91.3/-                  & 91.8/-                    & \textbf{92.5/-}                & 89.3/72.1                & \textbf{90.1/73.8}            \\
MNLI-m/mm (Accuracy)                    & 84.6/83.4      & \textbf{86.1/85.5}                  & 86.6/-                     & \textbf{89.8/-}                      & 89.1/-                  & 86.7/85.9                     & \textbf{88.7/88.8}                 \\
QNLI (Accuracy)                    & 90.5      & \textbf{92.9}                  & 92.3                     & 93.9                      & \textbf{94.3}                  & 92.7                     & \textbf{94.6}                 \\
RTE (Accuracy)                       & 66.4      & \textbf{74.8}                  & 70.4                     & 83.8                      & \textbf{85.2}                  & 70.1                     & \textbf{80.2}                 \\
WNLI (Accuracy)                      & \textbf{65.1}      & \textbf{65.1}                  & -                        & -                         & -                              & 65.1                     & \textbf{67.8}                 \\ 
AX(Matthew Corr.)           & 34.2  & \textbf{37.4}  &   - & - & -  & 39.6 & \textbf{48.0}    \\    \hline         
Score                                & 78.3      & \textbf{80.6}                  & -                        & -                         & -                              & 80.5                     & \textbf{83.6}                 \\ \hline \hline
\end{tabular}
\caption{The results on GLUE benchmark, where the results on dev set are the median of five runs and the results on test set are scored by the GLUE evaluation server (\url{https://gluebenchmark.com/leaderboard}). The state-of-the-art results are in bold. All of the fine-tuned models of AX is trained by the data of MNLI.}
\label{glue_test}
\end{table*}
\begin{table*}[htbp]
  \centering
  \small
\begin{tabular}{c|c|cc|cc|cc|cc}
\hline \hline
\multirow{2}{*}{Task} & \multirow{2}{*}{Metrics} & \multicolumn{2}{c|}{BERT$_{BASE}$} & \multicolumn{2}{c|}{ERNIE 1.0$_{BASE}$} & \multicolumn{2}{c|}{ERNIE 2.0$_{BASE}$} & \multicolumn{2}{c}{ERNIE 2.0$_{LARGE}$} \\ \cline{3-10} 
                      &                          & Dev         & Test        & Dev               & Test             & Dev                 & Test               & Dev                 & Test                \\ \hline
CMRC 2018             & EM/F1                    & 66.3/85.9   & -           & 65.1/85.1         & -                & 69.1/88.6           & -                  & \textbf{71.5/89.9}  & -                   \\
DRCD                  & EM/F1                    & 85.7/91.6   & 84.9/90.9   & 84.6/90.9         & 84.0/90.5        & 88.5/93.8           & 88.0/93.4          & \textbf{89.7/94.7}  & \textbf{89.0/94.2}  \\
DuReader              & EM/F1                    & 59.5/73.1   & -           & 57.9/72.1         & -                & 61.3/74.9           & -                  & \textbf{64.2/77.3}  & -                   \\
MSRA-NER              & F1                       & 94.0        & 92.6        & 95.0              & 93.8             & 95.2                & 93.8               & \textbf{96.3}       & \textbf{95.0}       \\
XNLI                  & Accuracy                 & 78.1        & 77.2        & 79.9              & 78.4             & 81.2                & 79.7               & \textbf{82.6}       & \textbf{81.0}       \\
ChnSentiCorp          & Accuracy                 & 94.6        & 94.3       & 95.2              & 95.4             & 95.7                & 95.5               & \textbf{96.1}       & \textbf{95.8}       \\
LCQMC                 & Accuracy                 & 88.8        & 87.0        & 89.7              & 87.4             & \textbf{90.9}       & \textbf{87.9}      & \textbf{90.9}       & \textbf{87.9}       \\
BQ Corpus             & Accuracy                 & 85.9        & 84.8        & 86.1              & 84.8             & 86.4                & 85.0               & \textbf{86.5}       & \textbf{85.2}       \\
NLPCC-DBQA            & MRR/F1                   & 94.7/80.7   & 94.6/80.8   & 95.0/82.3         & 95.1/82.7        & 95.7/84.7           & 95.7/85.3          & \textbf{95.9/85.3}           & \textbf{95.8/85.8}           \\ \hline \hline

\end{tabular}
\caption{The results of 9 common Chinese NLP tasks. ERNIE 1.0 indicates model released by \protect\cite[ERNIE]{sun2019ernie} . The reported results are the average of five experimental results, and the state-of-the-art results are in bold.}
  \label{finetune_table}
\end{table*}
\begin{table*}[htbp]
\tabcolsep 0.06in
  \centering
  \small
\begin{tabular}{c|c|cccc|ccc}
\hline \hline
\multirow{2}{*}{Pre-training method} & \multirow{2}{*}{Pre-training task} & \multicolumn{4}{c|}{Training iterations (steps) } &  \multicolumn{3}{c}{Fine-tuning result} \\ \cline{3-9} 
& & Stage 1 & Stage 2 & Stage 3 & Stage 4 & MNLI & SST-2 & MRPC \\ \midrule[1pt]

\multirow{4}{*}{Continual Learning} & Knowledge Masking & 50k & - & - & - & \multirow{4}{*}{77.3} & \multirow{4}{*}{86.4} & \multirow{4}{*}{82.5}\\ \cline{2-6}
                                        & Capital Prediction & - & 50k & - & - & \\  \cline{2-6}
                                        & Token-Document Relation  & - & - & 50k & - & \\  \cline{2-6}
                                        & Sentence Reordering  & - & - & - & 50k &\\\midrule[1pt]
\multirow{4}{*}{Multi-task Learning}  & Knowledge Masking & \multicolumn{4}{c|}{50k} & \multirow{4}{*}{78.7} & \multirow{4}{*}{87.5} & \multirow{4}{*}{83.0}\\  \cline{2-6}
                                        & Capital Prediction & \multicolumn{4}{c|}{50k} &  \\  \cline{2-6}
                                        & Token-Document Relation  & \multicolumn{4}{c|}{50k} &\\  \cline{2-6}
                                        & Sentence Reordering  & \multicolumn{4}{c|}{50k} & \\ \midrule[1pt]
\multirow{4}{*}{continual Multi-task Learning}  & Knowledge Masking & 20k & 10k & 10k & 10k & \multirow{4}{*}{\bf{79.0}}
& \multirow{4}{*}{\bf{87.8}} & \multirow{4}{*}{\bf{84.0}}\\  \cline{2-6}
                                        & Capital Prediction & - & 30k & 10k & 10k & \\  \cline{2-6}
                                        & Token-Document Relation & - & - & 40k & 10k & \\  \cline{2-6}
                                        & Sentence Reordering  & - & - & - & 50k & \\ 
\hline\hline

\end{tabular}
\caption{The results of different methods of continual pre-training. We use knowledge masking, capital prediction, token-document relation and sentence reordering as our pre-training tasks. we sample 10\% training data from our whole pre-training corpus. We train the model with 4 tasks altogether from scratch in multi-task learning method and train the model in 4 stages in other two learning methods. We train different tasks in different stages. The learning order of these tasks is the same as the above tasks listed. To compare the result fairly, each of these 4 tasks are updated in 50,000 steps . The size of pre-training model is same as ERNIE base. We choose MNLI-m, SST-2 and MRPC as our fine-tuning dataset. The fine-tuning result is average of five random start. the fine-tuning experiment set is same as Table \ref{GLUE_finetune_setting}.}
  \label{continual_learning}
\end{table*}
\subsection{Fine-tuning Tasks}

\subsubsection{English Task}
As a multi-task benchmark and analysis platform for natural language understanding, General Language Understanding Evaluation (GLUE) is usually applied to evaluate the performance of models. We also test the performance of ERNIE 2.0 on GLUE. Specifically, GLUE covers a diverse range of NLP datasets, the details is shown \cite{wang2018glue}.

\subsubsection{Chinese Tasks}
We executed extensive experiments on 9 Chinese NLP tasks, including machine reading comprehension, named entity recognition, natural language inference, semantic similarity, sentiment analysis and question answering. Specifically, the following Chinese datasets are chosen to evaluate the performance of ERNIE 2.0 on Chinese tasks:

\begin{itemize}
  \item \textbf{Machine Reading Comprehension (MRC)}: CMRC 2018 \cite{DBLP:journals/corr/abs-1810-07366}, DRCD \cite{shao2018drcd}, and DuReader \cite{he2017dureader}.
  \item \textbf{Named Entity Recognition (NER)}: MSRA-NER \cite{levow2006third}.
  \item \textbf{Natural Language Inference (NLI)}: XNLI \cite{conneau2018xnli}.
  \item \textbf{Sentiment Analysis (SA)}: ChnSentiCorp \footnote{https://github.com/pengming617/bert\_classification}.
  \item \textbf{Semantic Similarity (SS)}: LCQMC \cite{liu2018lcqmc}, and BQ Corpus \cite{chen2018bq}.
  \item \textbf{Question Answering (QA)}: NLPCC-DBQA \footnote{http://tcci.ccf.org.cn/conference/2016/dldoc/evagline2.pdf}.
\end{itemize}
\subsection{Implementation Details for Fine-tuning}
Detailed fine-tuning experimental settings of English tasks are shown in Table \ref{GLUE_finetune_setting} while that of Chinese tasks are shown in Table \ref{chinese_finetune_setting}.

\subsection{Experimental Results}
\subsubsection{Results on English Tasks}
We evaluate the performance of the base models and the large models of each method on GLUE. 
Considering the fact that only the results of the single model XLNet on the dev set are reported, we also reports the results of each method on the dev set. 
In order to obtain a fair comparison with BERT and XLNet, we run a single-task and single-model \footnote{which mean the result without additional tricks such as ensemble and multi-task losses.} ERNIE 2.0 on the dev set. The detailed results on GLUE are depicted in Table \ref{glue_test}. 

As shown in the \textit{BASE model} columns of Table \ref{glue_test}, ERNIE 2.0$_\text{BASE}$ outperforms BERT$_\text{BASE}$ on all of the 10 tasks and obtains a score of 80.6. As shown in the dev columns of \textit{LARGE model} section in Table \ref{glue_test}, ERNIE 2.0$_\text{LARGE}$ consistently outperforms BERT$_\text{LARGE}$ and XLNet$_\text{LARGE}$ on most of the tasks except MNLI-m. Furthermore, as shown in the \textit{LARGE model} section in Table \ref{glue_test}, ERNIE 2.0$_\text{LARGE}$ outperforms BERT$_\text{LARGE}$ on all of the 10 tasks, which gets a score of 83.6 on the GLUE test set and achieves a 3.1\% improvement over the previous SOTA pre-training model BERT$_\text{LARGE}$. 


\subsubsection{Results on Chinese Tasks}
Table \ref{finetune_table} shows the performances on 9 classical Chinese NLP tasks. It can be seen that ERNIE 1.0$_\text{BASE}$ outperforms BERT$_\text{BASE}$ on XNLI, MSRA-NER, ChnSentiCorp, LCQMC and NLPCC-DBQA tasks, yet the performance is less ideal on the rest, which is caused by the difference in pre-training between the two methods. Specifically, the pre-training data of ERNIE 1.0$_\text{BASE}$ does not contain instances whose length exceeds 128, but BERT$_\text{BASE}$ is pre-trained with the instances whose length are 512. From the results, it can be also seen that the proposed ERNIE 2.0 makes further progress, which significantly outperforms BERT$_\text{BASE}$ on all of the nine tasks. Furthermore, we train a large version of ERNIE 2.0. ERNIE 2.0$_\text{LARGE}$ achieves the best performance and creates new state-of-the-art results on these Chinese NLP tasks. 

\subsection{Comparison of Different Learning Methods}
In order to analyze the effectiveness of the continual multi-task learning strategy adopted in our framework, we compare this method with two other methods as shown in figure \ref{diff_method}. Table \ref{continual_learning} describes the detailed information. For all the methods, we assume that the training iterations are the same for each task. In our settings, each task can be trained in 50k iterations, with 200k iterations for all of the tasks. As it can be seen, multi-task learning trains all the tasks in one stage, continual pre-training trains the tasks one by one, while our continual multi-task learning method can assign different iterations to each task in different training stages.
The experimental result shows that continual multi-task learning obtains the better performance on downstream tasks compared with the other two methods, without sacrificing any efficiency. The result also indicates that our pre-training method can trains the new tasks in a more effective and efficient way. Moreover, the comparison between continual multi-task learning, multi-task learning and traditional continual learning shows that the first two methods outperform the third one, which confirms our intuition that traditional continual learning tends to forget the knowledge it has learnt when there is only one new task involved each time. 

\section{Conclusion}
We proposed a continual pre-training framework named ERNIE 2.0, in which pre-training tasks can be incrementally built and learned through continual multi-task learning in a continual way. 
Based on the framework, we constructed several pre-training tasks covering different aspects of language and trained a new model called ERNIE 2.0 model which is more competent in language representation. 
ERNIE 2.0 was tested on the GLUE benchmarks and various Chinese tasks. It obtained significant improvements over BERT and XLNet. In the future, we will introduce more pre-training tasks to the ERNIE 2.0 framework to further improve the performance of the model. We will also investigate other sophisticated continual learning method in our framework.

\bibliographystyle{aaai}  
\bibliography{references}  

\end{document}